\documentclass[runningheads]{llncs}
\usepackage{graphicx}
\usepackage{booktabs}
\usepackage{multirow}
\usepackage{arydshln}
\usepackage{caption}
\usepackage{subcaption}
\usepackage{amsmath,amssymb}
\usepackage{hyperref}
\usepackage{musicography} 
\usepackage{float}
\usepackage{acronym}
\acrodef{OMR}{Optical Music Recognition}
\acrodef{CER}{Character Error Rate}
\acrodef{SER}{Symbol Error Rate}
\acrodef{LER}{Line Error Rate}
\acrodef{HTR}{Handwritten Text Recognition}
\acrodef{OCR}{Optical Character Recognition}
\acrodef{DU}{Document Understanding}
\acrodef{CTC}{Connectionist Temporal Classification}
\acrodef{SMT}{Sheet Music Transformer}
\acrodef{CNN}{Convolutional Neural Network}
\acrodef{Swin-T}{Sliding Windows Transformer}

\newcommand{\krn}{\textsc{kern}}
\newcommand{\GrandStaff}{\textsc{GrandStaff}}
\newcommand{\SMTCNN}{$\text{SMT}_{\text{CNN}}$}
\newcommand{\SMTSWT}{$\text{SMT}_{\text{SWIN}}$}
\newcommand{\SMTNXT}{$\text{SMT}_{\text{NexT}}$}

\begin{document}
\title{Sheet Music Transformer: End-To-End Optical Music Recognition Beyond Monophonic Transcription
%\thanks{This paper is part of the project I+D+i PID2020-118447RA-I00 (MultiScore), funded by MCIN/AEI/10.13039/501100011033.}
}
\author{Antonio Ríos-Vila\inst{1}\orcidID{0000-0002-7770-9726}\and
Jorge Calvo-Zaragoza\inst{1}\orcidID{0000-0003-3183-2232}\and
Thierry Paquet\inst{2}\orcidID{0000-0002-2044-7542}}
%\author{Author names
%remain hidden for double-blind review.}

%
\authorrunning{Ríos-Vila et al.}
\institute{Pattern Recognition and Artificial Intelligence Group, University of Alicante, Spain \email{\{arios, jcalvo\}@dlsi.ua.es} \and
LITIS Laboratory - EA 4108, Rouen University, France \email{thierry.paquet@litislab.eu} \\}
\maketitle              % typeset the header of the contribution
\begin{abstract}
State-of-the-art end-to-end \acf{OMR} has, to date, primarily been carried out using monophonic transcription techniques to handle complex score layouts, such as polyphony, often by resorting to simplifications or specific adaptations. Despite their efficacy, these approaches imply challenges related to scalability and limitations. This paper presents the \acf{SMT}, the first end-to-end OMR model designed to transcribe complex musical scores without relying solely on monophonic strategies. Our model employs a Transformer-based image-to-sequence framework that predicts score transcriptions in a standard digital music encoding format from input images. Our model has been tested on two polyphonic music datasets and has proven capable of handling these intricate music structures effectively. The experimental outcomes not only indicate the competence of the model, but also show that it is better than the state-of-the-art methods, thus contributing to advancements in end-to-end OMR transcription.

\end{abstract}

\section{Introduction}
Music is a valuable component of our cultural heritage, as it is a resource that enables an understanding of the social, cultural and artistic trends of each period of history. Most existing documents have been transmitted in the form of printed and handwritten documents. In the same way that \ac{OCR} and \ac{HTR} are successfully applied in order to extract content from text images, \ac{OMR} is the research area that produces models that automatically recognize and transcribe sheet music~\cite{Calvo-Zaragoza2019}. 

The progress made in \ac{OMR}, which initially depended on multi-stage workflows~\cite{Rebelo2012}, has diversified with the emergence of Deep Learning solutions. Two major approaches currently exist: object detection workflows \cite{Tugenner2024,Baro2022},---in which notes are individually detected and assembled into a digital document---and holistic workflows~\cite{Shi2016,CalvoZaragoza2019b,Baro2019,Torras2021,Alfaro-Contreras22}, also known as \textit{end-to-end} methods, in which systems directly generate a symbolic representation of a region of a given document. The latter paradigm dominates the current state of the art of other applications, such as the recognition of text, speech or mathematical formulae \cite{Chowdhury2018,Chiu2018,Zhang2017}, and has also proven successful in OMR.

Despite the successful results obtained, \ac{OMR} has, to date, found solutions that are applicable only to monophonic scores\footnote{Monophonic scores are pieces of music in which only one voice is present}. However, many non-monophonic music documents have not been dealt with by the literature concerning \ac{OMR}.

This need to cover all music documents signifies that end-to-end \ac{OMR} is now shifting toward new projects that were not, until recently, considered feasible. Recent literature focuses on applying OMR systems to complex scenarios, such as full-page documents \cite{Castellanos2020}, the simultaneous recognition of music and lyrics \cite{MartinezSevilla2023} and polyphonic scores \cite{Sachida2021,RiosVila2023}. However, most of these advances are \textit{ad-hoc} adaptations or pipeline-based solutions of the current state of the art for monophonic music transcription. There is no solution that currently goes beyond monophonic transcription, but rather adaptations that reduce the problem in order to make it close to a monophonic scenario and solve it using state-of-the-art methods. According to the recent works of \ac{HTR} \cite{Singh2021,Dhiaf2023} and \ac{DU} \cite{Coquenet2023b,Kim2022}, \ac{OMR} should seek to break this monophonic-dependent barrier. In this paper, we propose the \acf{SMT}, an image-to-sequence approach---based on autoregressive Transformers---that is able to transcribe music input images beyond monophony, without adaptations or specific preprocessing steps. 
In order to test this approach, we deal with the challenge of polyphonic music transcription, which is the most complex in \ac{OMR}, and compare it to other current state-of-the–art approaches. We specifically test it in two common scenarios regarding sheet music: pianoform scores and those for string quartets. Our experiments empirically demonstrate that our solution provides robust results that clearly outperform the state of the art. The scientific contributions of this paper can be summarized as follows:

\begin{itemize}
    \item We propose the \ac{SMT}, the first image-to-sequence-based approach for music transcription that is able to deal with transcripts beyond the monophonic level. In our experiments, we demonstrate that this approach performs better than current state-of-the-art solutions.
    \item We explore and analyze different configurations for feature extraction in order to produce a model that is better suited to complex music layouts.
    \item We create an adaptation of a well-known music dataset for end-to-end \ac{OMR} that goes beyond monophonic-level transcription.
\end{itemize}

The remainder of the paper is structured as follows: in Section \ref{sec:e2eomr}, we present the polyphonic transcription challenge in \ac{OMR}, explain the current state of the art and discuss its limitations. In Section \ref{sec:methodology}, we describe the end-to-end \ac{SMT} Neural Network proposed in this paper, while the experiments conducted are presented in Section \ref{sec:experiments}, along with the results, Section \ref{sec:results}, and analysis, Section \ref{sec:discussion}, that are drawn from them. Finally, we conclude the paper in Section \ref{sec:conclusions}, in addition to discussing other avenues for future work.
 
\section{End-to-end polyphonic OMR}
\label{sec:e2eomr}
The state-of-the-art holistic \ac{OMR} currently addresses music transcription as a sequence recognition problem, in which the most probable symbolic representation $\hat{\mathbf{s}}$---encoded in the $\Sigma_a$ music notation vocabulary---for each staff-section image $x$ is sought.

\begin{equation}
\label{eq:map_agn}
\hat{\mathbf{s}} = \arg\max_{\mathbf{s} \in \Sigma_{a}} P(\mathbf{s} \mid x)
\end{equation}

Neural network approaches approximate this probability by training with the \ac{CTC} loss~\cite{Graves2006}. The output of the network consists of a probability map, which contains the probabilities of all the tokens within the vocabulary $\Sigma_{a}$. In order to allow for the possibility of no prediction at a given timestep, CTC provides an extra blank token ($\epsilon$), and the output vocabulary of the network therefore becomes $\Sigma'_{a} = \Sigma_{a} \cup \{\epsilon\}$.

At inference, OMR methods employ a \textit{greedy} decoding, from which the most probable sequence is retrieved given an input image $x$. This formulation forces networks to define a reshape function based on the vertical collapse of the feature map, as a sequence must be retrieved. This vertical collapse assumes that the symbols can be read from left to right and that the frames\footnote{Column-wise elements of the image} of the feature map always contain information about the same symbol in this case.

\subsection{The challenge of transcribing polyphonic scores}
This methodology, which is able to address single-staff music transcription, has proven successful for the transcription of both printed and handwritten scores~\cite{Torras2021,Baro2019,CalvoZaragoza2019b,Calvo-Zaragoza2018} and has attained very accurate results, surpassing those of any other image-to-sequence approach~\cite{RiosVila2022}.

Despite this, the approach is severely limited. As mentioned in Section \ref{sec:e2eomr}, the approach performs a vertical collapse to group all the information regarding the image into a sequential structure, which simplifies the problem. However, this also prevents the network from dealing with multiple structures, i.e., music staves or symbols. This entails a difficulty when transcribing several music structures, even with the simplest graphical complexity. 

If we consider scores other than those of a monophonic nature, such as polyphonic scores, the challenge increases. These engravings must be read in a particular order during interpretation, since there are staves that must be read simultaneously. Rather than performing a line-by-line reading from top to bottom and left to right, interpretation is tied to staff groups, namely systems, in which all elements are read simultaneously from left to right (see Figure \ref{fig:kernsample}).

This particularity poses several challenges, since the premise that a frame contains the information regarding a single music symbol---even at the staff level---no longer holds in this case. Some vocabulary-based shortcuts can be taken if complexity does not increase much, as occurs in the work of Alfaro-Contreras et al.~\cite{Alfaro-Contreras19}. These approaches could be extended to more complex scenarios, such as polyphonic scores, but at the cost of increasing the length of the ground truth sequence. This scalability issue becomes unfeasible with larger samples---e.g. scores for string quartets---since vertical collapse cannot produce sufficient frames with which to accurately transcribe the score. It would, therefore, appear that methodological advances must be made in order to produce more robust and generic systems.

\subsection{Current approaches}
\label{sec:sota}
The first avenue found in the literature concerns employing a \textit{divide and conquer} approach, in which single-staff images are segmented by means of a Layout Analysis method and transcribed using state-of-the-art methods, as occurs in~\cite{Sachida2021}. However, despite proposing a robust solution to transcription, the alignment retrieval---in which simultaneous music notes must be placed in the same timestep in the digital document---of the music score becomes a challenging process, since the system retrieves individual decontextualized single-staff transcriptions. Exhaustive post-processing is, therefore, required in order to retrieve a correct and usable music document.

A holistic approach dealing with the aforementioned challenge recently appeared in~\cite{RiosVila2023}. Its authors specifically suggest an extension to the current state-of-the-art-based musical score alignment by exploiting music encoding features and score rotations. This method consists of a neural network that unfolds the rotated input score image and sequentially transcribes all simultaneous events as text lines in the ground truth---with special tokens that indicate polyphonic timesteps and the conclusion of simultaneous events. This system was tested on excerpts from single-line pianoform music, which are the most complex in \ac{OMR}\footnote{The reason for the difficulty related to pianoform scores is that they contain multiple voices within a single staff and cross-staff interactions that are not easily seen in the image of the document.}. Although the aforementioned paper presents very competitive results in the case of excerpts from single-system pianoform music, the approach is very limited. Indeed, since this model unfolds the score in order to detect simultaneous events, this system is still limited to input images in which all the events are simultaneous, and does not allow multiple systems in a single image. A complementary Layout Analysis should, therefore, be performed in order to address complex documents. Moreover, this method has not been stressed in larger images with simultaneous events, with scalability being another potential issue when using this system.

\section{Sheet Music Transformer}
\label{sec:methodology}

\begin{figure}[H]
\centering
    \includegraphics[width=\textwidth]{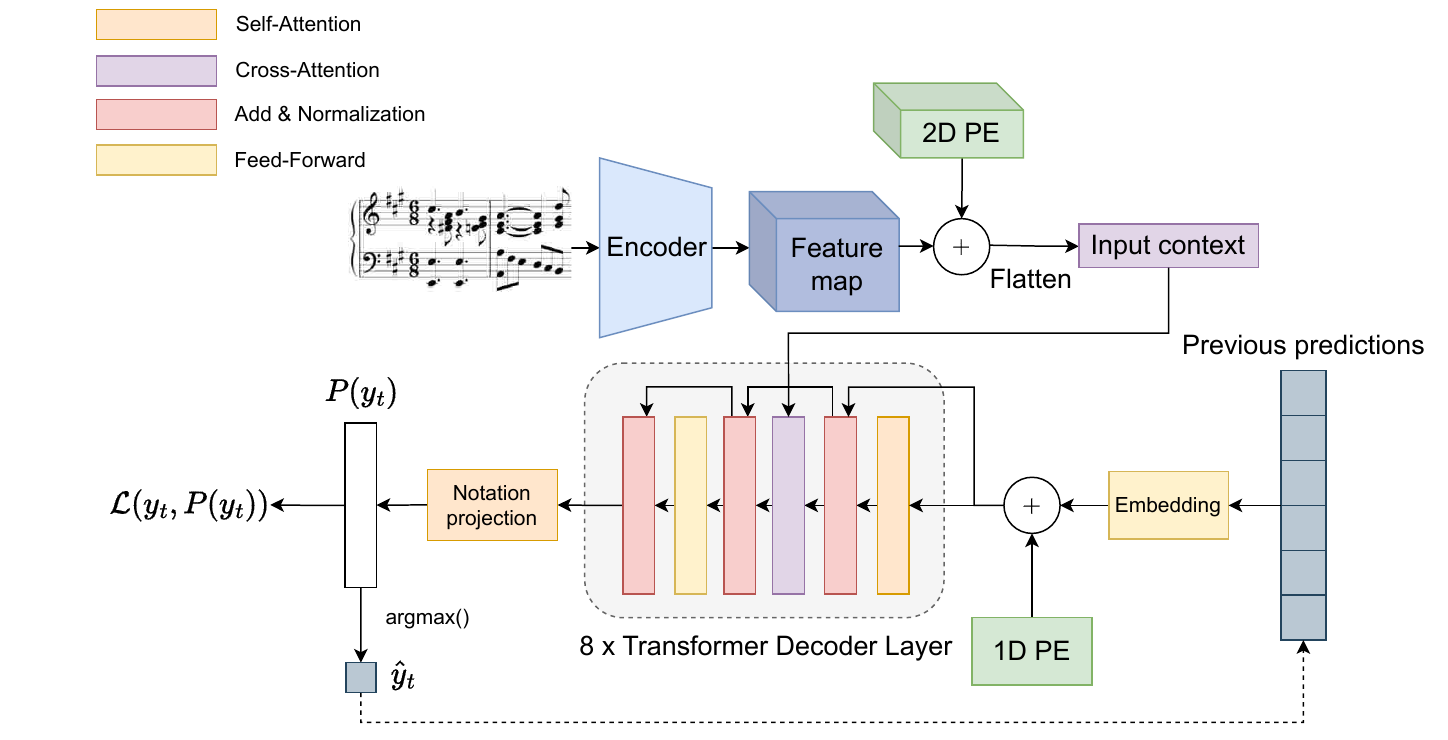}
    \caption{Graphic scheme of the \ac{SMT} architecture.}
    \label{fig:SMT}
\end{figure}

In this paper, we present the Sheet Music Transformer (SMT). This model is an autoregressive end-to-end neural network that generates the transcription of a given polyphonic music system input image. The model consists of two fundamental components: an encoder and a decoder. The encoder acts as a feature extractor of the image $x$, producing a feature map $x_e^{\prime}$. The decoder consists of an autoregressive conditioned language model that predicts the probability of each symbol of the vocabulary in a timestep given the output feature vector from the encoder and the previously generated symbols. This is formalized as: 
\begin{equation}
    \hat{y} \underset{\hat{y} \in \Sigma}{=} P(\hat{y}_t \mid x_e^{\prime}, \left(\hat{y}_0, \hat{y}_1, \hat{y}_2, ..., \hat{y}_{t-1}\right))
\end{equation}
where $\Sigma$ represents the symbol vocabulary used in order to encode the music content, $x$ is the input feature map and $t$ is the current timestep. A graphic scheme of the \ac{SMT} is depicted in Figure \ref{fig:SMT}.

\subsection{Encoder}

Let $x \in \mathbb{R}^{c \times h \times w}$ represent the input image of the system, where $h$ and $w$ respectively denote its height and width in pixels and $c$ denotes the number of channels. This block is typically implemented through the use of Convolutional Neural Networks~\cite{Coquenet2023b,Singh2021}, owing to their capacity to process image signals. This module therefore results in an image processor that outputs $c_e$ two-dimensional feature maps denoted by $x_e \in \mathbb{R}^{h_e \times w_e \times c_e}$. Note that, $h_e$ and $w_e$ respectively relate to the image dimensions $h_e = \frac{h}{r_h}$ and $w_e = \frac{w}{w_t}$, where $r_h$ and $r_w$ represent the corresponding downscaling produced by this network.

\subsection{Decoder}

In this work, we have opted for a Transformer~\cite{Vaswani2017} decoder, as it is currently the state-of-the-art method for tasks involving conditional sequence generation of varying lengths owing to its capacities to capture temporal relationships within the data through the use of multi-head attention mechanisms. 

The decoder consists of a language model that, at each time step $t$ (which is conditioned by the input feature vector produced by the encoder, $x_e^{\prime}$ along with the previously predicted tokens, $\left(\hat{y}_0, \ldots, \hat{y}_{t-1}\right)$) outputs a probability distribution $p_t \in \mathbb{R}^{\left|\Sigma\right|}$ over the $\Sigma$ vocabulary symbols, with the predicted token $\hat{y}_t$ being that which maximizes this probability score. The process starts with a special start-of-transcription symbol, $\hat{y}_0 = \text{\textless \textit{sot}\textgreater}$, and continues until an end-of-transcription token is predicted, $\hat{y}_{|\mathbf{\hat{y}}|} = \text{\textless \textit{eot}\textgreater}$.

We should stress that the encoder outputs a two-dimensional feature map to serve as a context in which to condition the decoder. However, the decoder works with one less dimension, since it is a sequence generator. In order to connect these two modules, the two-dimensional feature maps must be converted into a one-dimensional format that is suitable for the decoder, with a straightforward approach being that of flattening this structure across height and width, resulting in sequences of length $h_e \times w_e$.

Another aspect to take into account is that the Transformer implements a positional encoding (PE) mechanism with which to model its otherwise order-agnostic operation. This mechanism adds a position vector to each input element, determined by its location in the sequence. The addition of a one-dimensional PE to the unfolded feature maps could initially be considered possible, but while this might be sufficient for monophonic music sequences, it would result in a loss of spatial information when targeting polyphonic scores in which multiple voices are played simultaneously. In order to ensure that the model is aware of all the dimensions of the image, we, therefore, incorporate a two-dimensional PE within the feature maps before they are flattened into a one-dimensional sequence. The two-dimensional PE proposal is based on sine and cosine functions, akin to the original one-dimensional PE, in which the first half of the feature dimensions---i.e., $[0, c_e/2)$---is meant for horizontal positions, whereas the second half---i.e., $[c_e/2, c_e)$---is used for the vertical ones, the same that in~\cite{Singh2021,Coquenet2023b}. %This encoding is described in Eq.~\ref{eq:PE2D}:

\begin{align}
\begin{split}
\label{eq:PE2D}
    \text{PE}_{\text{2D}}\left(\text{pos}_t, 2i\right) & =\sin \left(\text{pos}_t/10000^{2i / c_e}\right) \\
    \text{PE}_{\text{2D}}\left(\text{pos}_t, 2i + 1\right) & =\cos \left(\text{pos}_t/10000^{2i / c_e}\right) \\
    \text{PE}_{\text{2D}}\left(\text{pos}_f, c_e/2 + 2i\right) & = \sin \left(\text{pos}_f/10000^{2i / c_e}\right) \\
    \text{PE}_{\text{2D}}\left(\text{pos}_f, c_e/2 + 2i + 1\right) & = \cos \left(\text{pos}_f/10000^{2i / c_e}\right)
\end{split}
\end{align}
where $\text{pos}_t$ and $\text{pos}_f$ respectively specify the horizontal (width) and vertical (height) pixel positions and $i \in\left[0, c_e / 4\right)$ denotes the feature dimension of the output map.

\subsection{Output Encoding}
The definition of an output encoding for our system is another key step to cover. Of the variety that OMR provides, the first choices to consider are the most widespread musical encodings in digital musicology: MEI~\cite{Hankinson2011} and MusicXML~\cite{Good2003}, which represent the components and metadata of a musical score in an XML-based markup encoding. Despite their extensive capabilities, these formats are overly verbose, which is not convenient for \ac{OMR} systems, 
as their conversion to deep-learning-friendly formats tends to be lossy.

\begin{figure}[ht]
    \centering
    \includegraphics[width=0.8\textwidth]{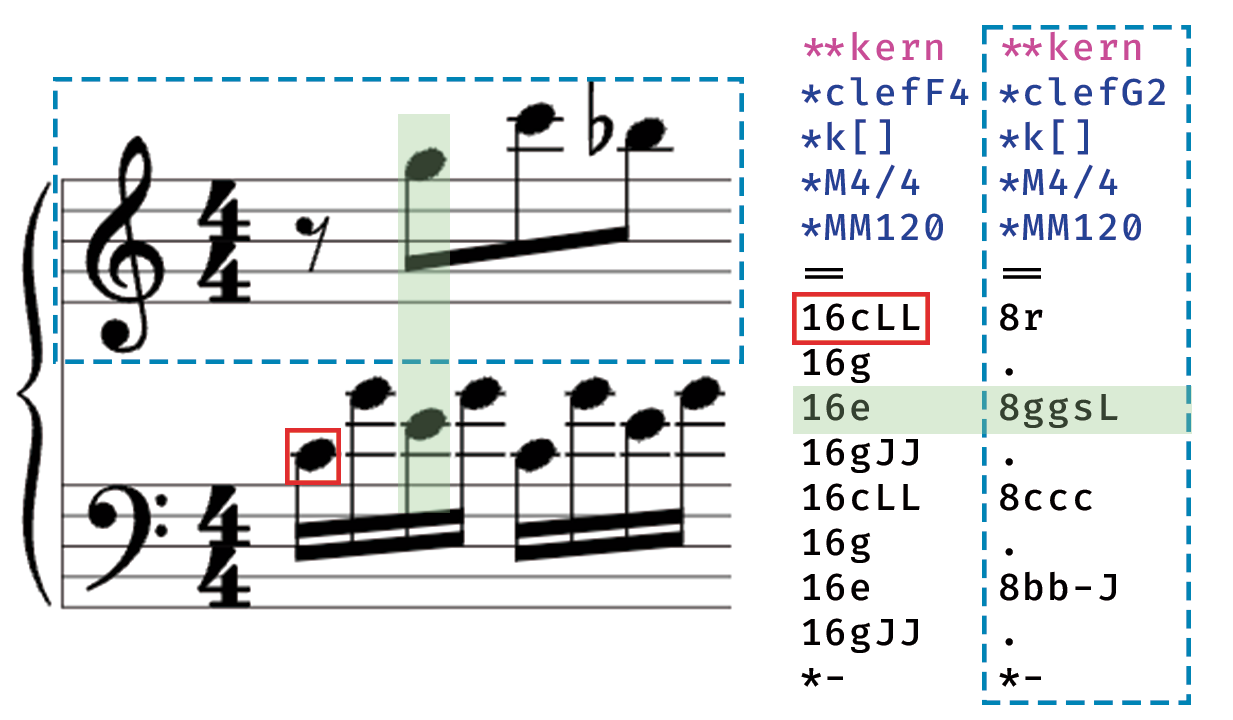}
    \caption{Example of an excerpt from Humdrum **kern-encoded pianoform music. The blue dashed line represents a **kern spine, which is a voice in the musical score, in this case, a staff. The green box represents simultaneous notes during interpretation, which are represented as a text line in the ground truth. The red box is a single musical symbol (in this case, a note). The ground truth is read top to bottom and left to right, while the score is read from left to right and bottom to top.}
    \label{fig:kernsample}
\end{figure}

In this paper, we have employed the text-based Humdrum **kern encoding format, which is included in the Humdrum tool-set~\cite{Huron1997} and is hereafter referred to simply as \krn{}. This music notation is one of the representations found most frequently in computational music analysis. Its features include a simple vocabulary and an easy-to-parse file structure, which is highly suitable for end-to-end \ac{OMR} applications. Moreover, \krn{} files are compatible with software dedicated to music~\cite{Sapp2017,Pugin2014} and can be automatically converted to other music encodings, such as those mentioned above, by means of straightforward operations. A simple example of a Humdrum **kern-encoded score\footnote{Please refer to the official Humdrum **kern syntax (\url{https://www.humdrum.org/rep/kern/}) for a more detailed explanation.} is shown in Figure \ref{fig:kernsample}.

\section{Experimental setup}
\label{sec:experiments}

In this section, we present the experimental framework designed in order to evaluate the performance of our method in comparison to other state-of-the-art methods\footnote{The implementation of the model and the datasets are available in \url{https://grfia.dlsi.ua.es/sheet-music-transformer/}}.

\subsection{Corpora}

Two different datasets of polyphonic music scores have been used in order to perform our experiments.

\begin{figure}[H]
     \centering
     \begin{subfigure}[b]{0.8\textwidth}
         \includegraphics[width=\textwidth]{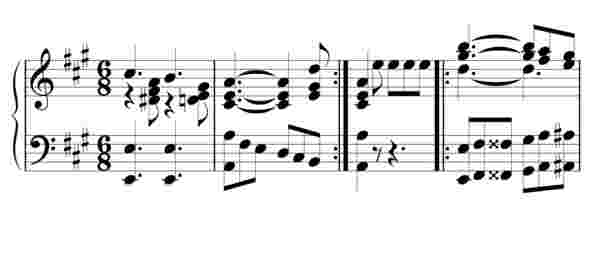}
         \caption{\GrandStaff{} music excerpt}
         \label{subfig:grandstaff_sample}
     \end{subfigure}
     \begin{subfigure}[b]{0.8\textwidth}
        \includegraphics[width=\textwidth]{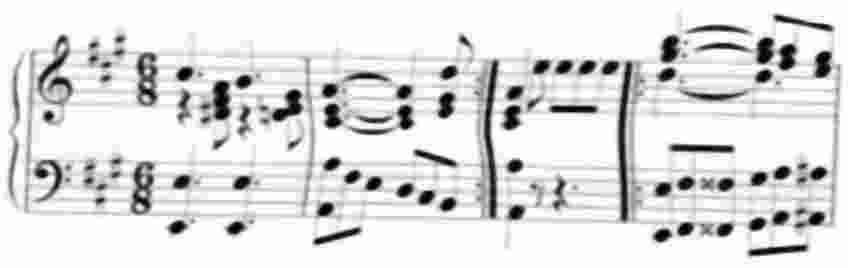}
         \caption{Distorted version of (a).}
         \label{subfig:camera_grandstaff_sample}
     \end{subfigure}
        \caption{Examples of the data contained in the \GrandStaff{} and Camera \GrandStaff{} dataset.}
        \label{fig:grandstaff}
\end{figure}

\subsubsection{GrandStaff} 
The \GrandStaff{} dataset is a publicly available corpus\footnote{\url{https://sites.google.com/view/multiscore-project/datasets}}~\cite{RiosVila2023} that consists of $53,882$ printed images of single-line (or system) pianoform scores, along with their digital score encoding. The dataset is composed of both original works from six authors---from the Humdrum\footnote{\url{https://github.com/humdrum-tools/humdrum-data}} repository---and synthetic augmentations of the music encodings that make it possible to provide a greater variety of musical sequences and patterns. The dataset comes with an official partition, in which the $7,000$ original scores are used as a test set, and the $46,882$ samples generated from augmentations comprise the training set. The dataset comes with an alternative version that introduces distortions into images to make them resemble low-quality photocopies. This version is, from here on, referred to as Camera \GrandStaff{}. Figure \ref{fig:grandstaff} depicts two samples of this dataset.

\subsubsection{Quartets}
\begin{figure}
    \centering
    \includegraphics[width=\textwidth]{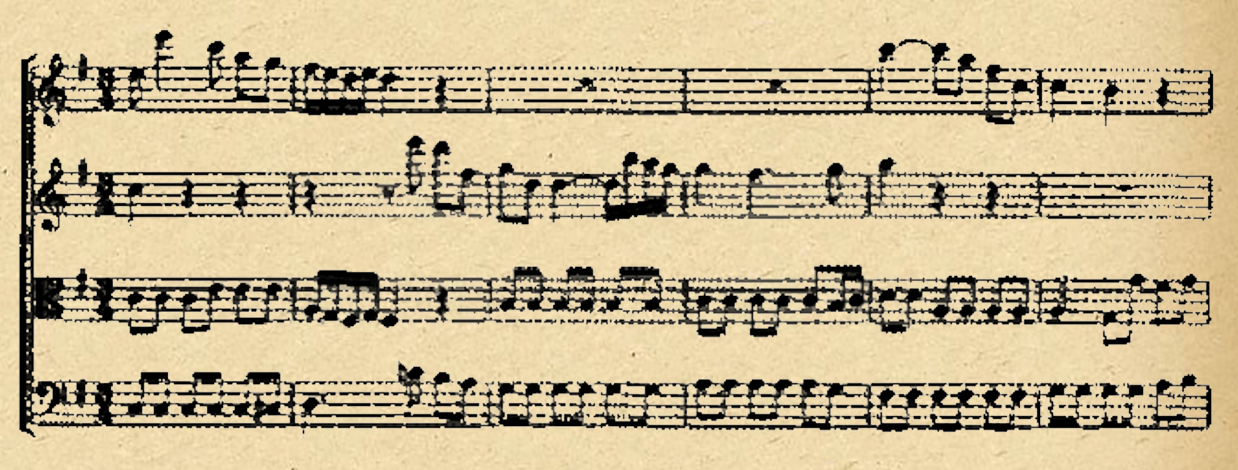}
    \caption{Examples of an excerpt of music from the Quartets dataset.}
    \label{fig:quartets_example}
\end{figure}

Since the \GrandStaff{} dataset is the only publicly available corpus designed for polyphonic end-to-end OMR, we produced an additional dataset with which to test both the state-of-the–art methods and our approach in a different polyphonic scenario. In this paper, we introduce the Quartets dataset. Quartets is a well-known collection employed in the Audio to Score field~\cite{Roman2019,Arroyo2022}. As the dataset provides the Humdrum **kern transcriptions from the excerpts of music, we produced a single-system transcription version of it. The dataset provides pieces that were randomly split from the original audios, namely pieces, into portions of approximately seven seconds, resulting in a total of $38\,051$ excerpts. These excerpts were rendered into printed music images using the Verovio Tool~\cite{Pugin2014}. Once the music images had been generated, we distorted the image using the same operations as those employed with Camera \GrandStaff{}\footnote{All of the operations pertain to the ImageMagick image processing library.} and included an additional distortion that simulated old printed ink, which contains bleeding and erasing errors. This distorted image was eventually fused with a random texture from a set of images on old paper. An example of this dataset is depicted in Figure \ref{fig:quartets_example}.

We followed the same partitions as those provided in the original dataset. The training set specifically contains $18\,162$ samples for Haydn, $7\,435$ samples for Mozart, and $12\,454$ samples for Beethoven. Each corpus is divided into three splits at piece level: train (70\%), validation (15\%), and test (15\%), and are combined in order to retrieve the partitions of the corpus. Please note that the partitions are made at the piece, level, meaning that excerpts originating from the same score will always be in the same partition, thus avoiding potential test biases.

\subsection{Neural networks configuration}
This paper introduces an image-to-sequence approach whose objective is to transcribe music scores that are more complex than monophony.

\subsubsection{Baseline} Section \ref{sec:sota} mentions the work of \cite{RiosVila2023}, which is currently the only end-to-end approach that transcribes musical scores other than those of a monophonic nature, with some \textit{ad-hoc} adaptations. We propose this system as our baseline, since its results were obtained using the \GrandStaff{} dataset and can easily be applied to that of the Quartets. We first implemented the best performing version of the model the authors presented with the raw Humdrum **kern notation, which is a Fully Connected Convolutional Neural Network, as the encoder, with a Recurrent Neural Network as a decoder, namely CRNN. In order to provide a fair comparison---which has not been explored in state of the art for the Quartets dataset---we also implemented the Transformer decoder version that the authors presented in the paper, which is referred to as CNNT\footnote{The baseline experiments were performed with the public implementation of the model, which can be found at \url{https://github.com/multiscore/e2e-pianoform}}.

\subsubsection{SMT}

We propose an image-to-sequence approach with three different feature extractors. The common aspect of all the \ac{SMT} tested in this paper is the implementation of a decoder, and we specifically follow the implementation of~\cite{Coquenet2023b}. The \ac{SMT} contains a transformer decoder with eight layers, four attention heads and an embedding size of 256 features in both the attention and feed-forward modules. Given this common framework, the three variants of the feature extractor are the following:

\begin{itemize}
    \item \textbf{CNN} As occurs in the recent works of \ac{HTR}~\cite{Singh2021,Coquenet2023b,Dhiaf2023}, the feature extractor is a fully-connected \ac{CNN}. We specifically implement the same backbone as that employed in ~\cite{Coquenet2023b}, as it is the lightest network that can achieve good results. This convolutional backbone is updated in order to rescale the height of the image by 16, since this attained the best results in previous experimentation. This implementation is referred to as \SMTCNN{}. 
    \item \textbf{Swin Transformer} The results of~\cite{Kim2022} inspired us to consider the \ac{Swin-T} model as our feature extractor. This model is an adaptation of the Vision Transformer~\cite{Dosovitskiy2021} with sliding-window mechanisms and patch-merging layers that imitate the hierarchical behavior of well-known \ac{CNN} backbones. In order to attain comparable results to those of the \SMTCNN{}, our \ac{Swin-T} implementation maintains only the first three layers from the network, with feature spaces of $64$, $128$ and $256$, maintaining its original layer ratio. This implementation is referred to as \SMTSWT{}. 
    \item \textbf{ConvNexT} Given the performance results of the \ac{Swin-T} in the Computer Vision field, the work by Liu et. al.~\cite{Liu2022} proposes an adaptation of traditional \ac{CNN} backbones, specifically the ResNet50, so as to imitate the behavior of the \ac{Swin-T}. For the sake of completion, we implement an adaptation of the ConvNexT as a backbone for our network. We specifically maintain only the first three layers---with the same ratios---of the network with depths of $64$, $128$ and $256$.
\end{itemize}

\subsection{Metrics}
The assessment of the performance of OMR experiments is one of the main challenges of the field. Despite several efforts~\cite{Torras2023,Hajic2017}, OMR does not provide a standard evaluation framework. In the case of end-to-end approaches, it is convenient to employ text-related metrics in order to perform this evaluation. Since we are comparing our method to current state-of-the-art approaches in this paper, we employ the same metrics as those presented in \cite{RiosVila2023}.  All of these measures are based on the normalized mean edit distance between a hypothesis sequence and a reference sequence by the length of the reference. These measures are the \ac{CER}, which evaluates among minimum semantic units in the output encoding, \ac{SER}, a commonly-used metric that takes complete symbols as evaluation tokens, and \ac{LER}, which approximates the accuracy of the model as regards both content and **kern document structure retrieval.\footnote{Evaluating the structure of the document is a key aspect in OMR when the transcription is more complex than the staff level, since the correct alignment between transcribed notes is relevant for a retrieval of the correct interpretation}

\section{Results}
\label{sec:results}
Table \ref{tab:results} shows the results attained for both the baseline and our approach for the \GrandStaff{}, Camera \GrandStaff{} and Quartets collections. 

\begin{table}[]
\centering
\renewcommand{\arraystretch}{1.1}
\caption{Average CER, SER, and LER (\%) obtained by the models studied in the test set for both the perfectly printed and the distorted versions of the corpora presented in this work.}
\label{tab:results}
\begin{tabular}{llccclccclccc}
\hline
\multicolumn{1}{c}{\textbf{Method}} &  & \multicolumn{7}{c}{\textbf{GrandStaff}} &  & \multicolumn{3}{c}{\multirow{2}{*}{\textbf{Quartets}}} \\
 &  & \multicolumn{3}{c}{\textbf{Ideal}} &  & \multicolumn{3}{c}{\textbf{Camera}} &  & \multicolumn{3}{c}{} \\
\multicolumn{1}{c}{\textbf{}} &  & \textbf{CER} & \textbf{SER} & \textbf{LER} &  & \multicolumn{1}{l}{\textbf{CER}} & \textbf{SER} & \textbf{LER} &  & \textbf{CER} & \textbf{SER} & \textbf{LER} \\ \hline
\multicolumn{13}{l}{\textit{Baseline}} \\
~ ~ CRNN & \multicolumn{1}{c}{} & 5.0 & 7.3 & 23.2 & \multicolumn{1}{c}{} & 7.2 & 9.9 & 29.5 &  & 21.9 & 22.5 & 67.0 \\
~ ~ CNNT & \multicolumn{1}{c}{} & 7.9 & 11.1 & 32.4 & \multicolumn{1}{c}{} & 9.4 & 12.3 & 33.3 &  & 30.8 & 32.6 & 82.4 \\ \hdashline
\multicolumn{13}{l}{\textit{Approach}} \\
~ ~ \SMTCNN{} &  & 5.7 & 7.6 & 19.5 &  & 6.9 & 8.5 & 20.2 & \multicolumn{1}{c}{} & 2.8 & 3.0 & 10.9 \\
~ ~ \SMTSWT{} &  & 53.2 & 70.1 & 98.2 & \multicolumn{1}{c}{} & 60.3 & 82.3 & 100.0 & \multicolumn{1}{c}{} & 40.5 & 50.2 & 75.7 \\
~ ~ \SMTNXT{} &  & \textbf{3.9} & \textbf{5.1} & \textbf{13.1} &  & \textbf{5.3} & \textbf{6.2} & \textbf{13.5} & \multicolumn{1}{c}{} & \textbf{1.3} & \textbf{1.4} & \textbf{5.6} \\ \hline
\end{tabular}
\end{table}

We first demonstrate that the state-of-the-art method struggles when the number of voices increases in the music score, with a \ac{CER} of 21.9\%, a \ac{SER} of 22.5\% and a \ac{LER} of 67.0\% in the Quartets Dataset, which are much worse than those ones for both \GrandStaff{} and Camera \GrandStaff{}. 
Given the results attained by our approach, we can clearly determine that the Convolution-based approaches, \SMTCNN{} and \SMTNXT{}, considerably outperform the state-of-the–art method, particularly as regards Camera \GrandStaff{} and Quartets. The \SMTSWT{}, however, attains the worst results. This underperformance can be explained by the fact that the Swin-T model, although light and powerful, is composed entirely of self-attention layers. Since the Transformer architecture has a high inductive bias and no pretraining is performed, it is reasonable to assume that this model requires a greater amount of samples than the \ac{CNN} and the ConvNexT in order to learn to extract relevant features from the image. Despite this, the \SMTNXT{} proposal attains outstanding results for all the datasets. There are improvements as regards \ac{CER}, \ac{SER} and \ac{LER} of 22\%, 32.9\% and 43.5\% in \GrandStaff{}; 26.4\%, 37.4\% and 54.2\% in Camera \GrandStaff{}, and 91.8\%, 91.6\% and 89.1\% in the Quartets dataset. These results demonstrate that our image-to-sequence approach is both a viable and scalable option with which to transcribe music more complex than monophony without \textit{ad-hoc} adaptations.

\section{Discussion}
\label{sec:discussion}
Although the results described in the previous section are positive, an analysis of the \ac{SMT} performance is required in order to explain and understand these improvements. This study is carried using the model that performed best in our experiments: the \SMTNXT{}.

One of the most noteworthy results is the substantial improvement as regards the \ac{LER} metric. For example, the \SMTCNN{}, despite producing slightly worse \ac{CER} and \ac{SER} than the CRNN baseline, notably improves the \ac{LER}. Although this metric suggests an overall improvement to the output quality, it is difficult to assess what this implies. In order to clarify this, we evaluated what percentage of the test results had a correct **kern structure. That is, we assessed how many documents could be processed by a standard musicological tool, such as Verovio Humdrum Viewer~\cite{Pugin2014} or MuseScore. The results of this method are depicted in Table \ref{tab:quality_analysis}.

\begin{table}[]
\centering
\caption{Average Render percentage (\%) values for the best-performing models for both the baseline and our approach in the test sets for the collections evaluated.}
\label{tab:quality_analysis}
\begin{tabular}{@{}lclccccccc@{}}
\toprule
\multirow{2}{*}{} &
  \multicolumn{1}{l}{\multirow{2}{*}{\textbf{Samples}}} &
   &
  \multicolumn{3}{c}{\textbf{CRNN}} &
  \textbf{} &
  \multicolumn{3}{c}{\textbf{\SMTNXT{}}} \\
 &
  \multicolumn{1}{l}{} &
   &
  \multicolumn{1}{l}{\textbf{\ac{LER}}} &
  \multicolumn{1}{l}{} &
  \textbf{Render \%} &
  \multicolumn{1}{l}{} &
  \multicolumn{1}{l}{\textbf{\ac{LER}}} &
  \multicolumn{1}{l}{} &
  \textbf{Render \%} \\ \midrule
\GrandStaff{}        & 7661 &  & 23.2 &  & 70.3 &  & 13.1 &  & 98.6 \\
Camera \GrandStaff{} & 7661 &  & 29.5 &  & 51.1 &  & 13.5 &  & 96.4 \\
Quartets          & 6107 &  & 67.0 &  & 22.5 &  & 5.6  &  & 90.7 \\ \bottomrule
\end{tabular}
\end{table}
These results make it possible to state that there is a clear correlation between \ac{LER} and overall document quality, as an improvement to this metric significantly increases the number of directly usable documents, which can be directly edited and processed with musicological tools. It is consequently possible to state that the \ac{SMT} produces outputs that are easier for end-users to edit and handle, thus making them more usable.

\begin{figure}[H]
     \centering
     \begin{subfigure}[b]{0.8\textwidth}
         \includegraphics[width=\textwidth]{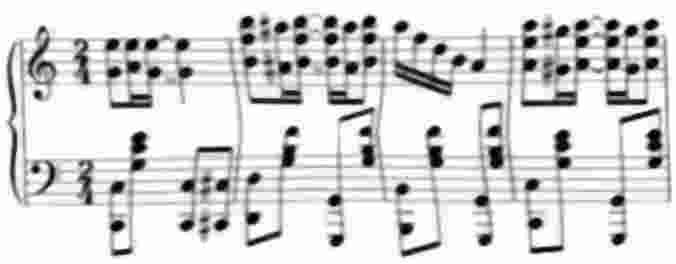}
         \caption{Input test image}
         \label{subfig:input_grandstaff}
     \end{subfigure}
     \hfill
     \begin{subfigure}[b]{0.8\textwidth}
        \includegraphics[width=\textwidth]{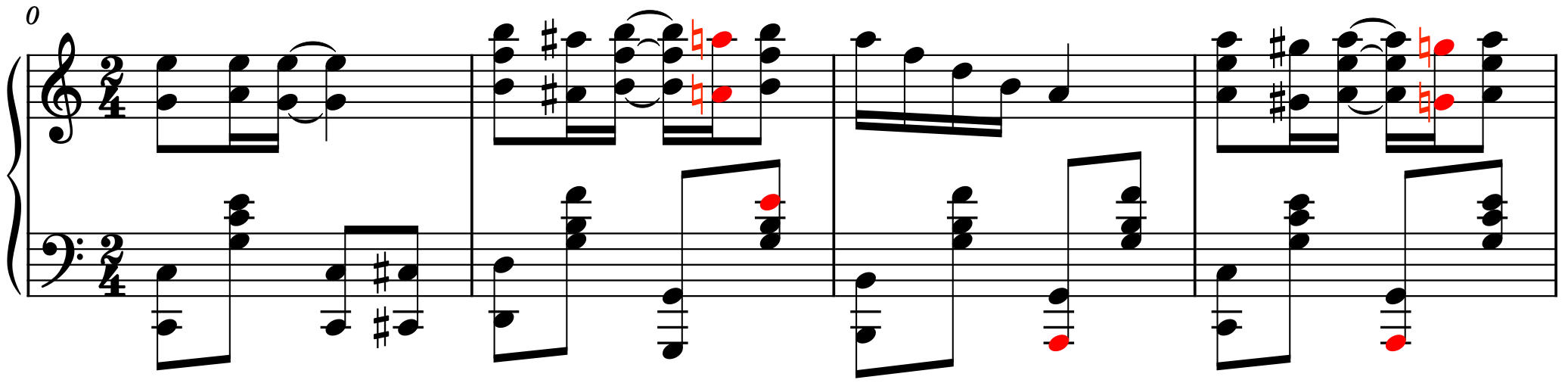}
         \caption{Render of the prediction made by the \SMTNXT{} with highlighted transcription errors.}
         \label{subfig:output_grandstaff}
     \end{subfigure}
        \caption{Test example from the \GrandStaff{} dataset with the errors highlighted. This specific sample attained a \ac{CER} of 5.0\%, a \ac{SER} of 6.0\% and a \ac{LER} of 20.5\%.}
        \label{fig:grandstaff_test_ex}
\end{figure}

\begin{figure}[ht]
     \centering
     \begin{subfigure}[b]{0.49\textwidth}
         \includegraphics[width=\textwidth]{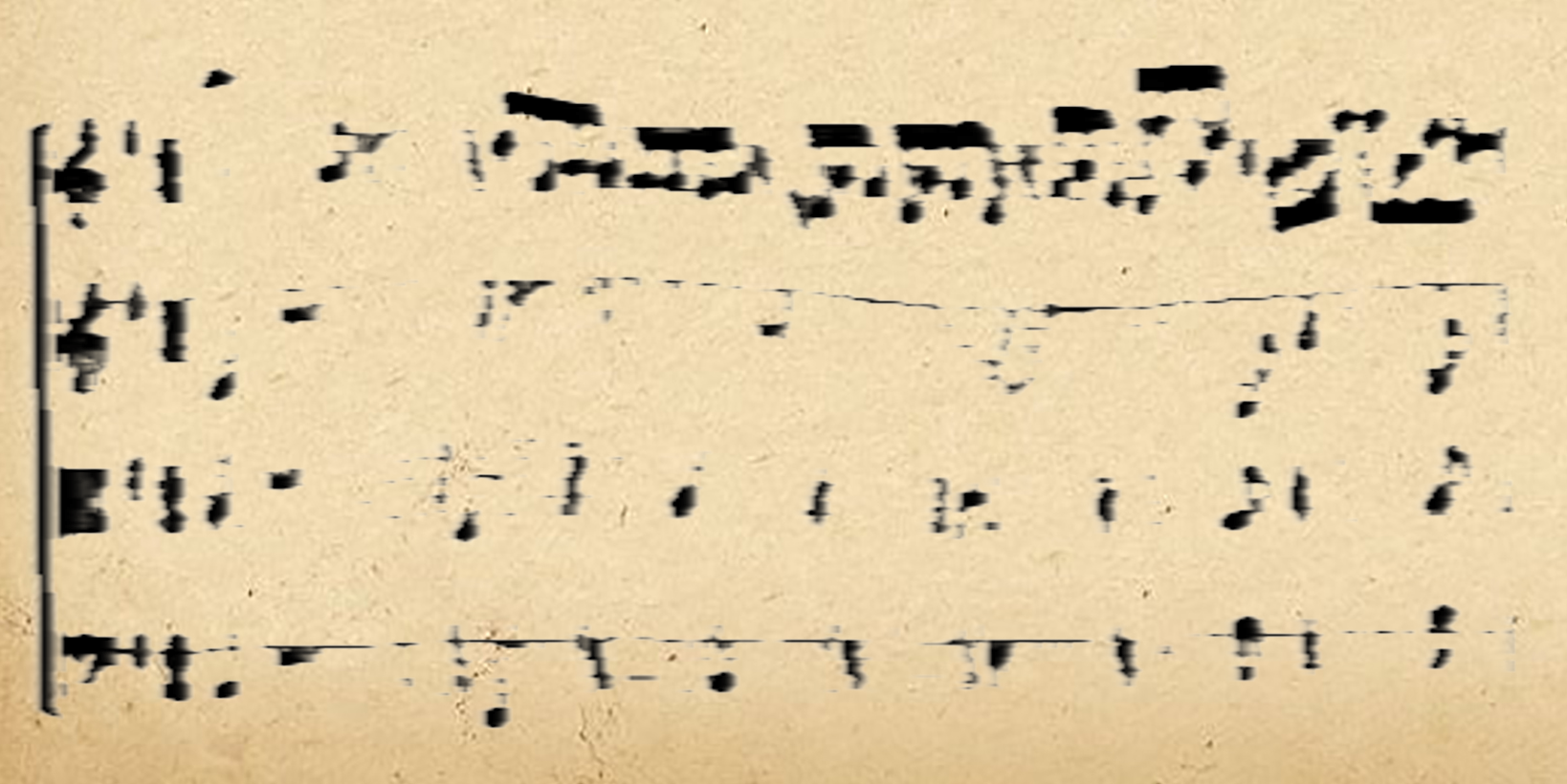}
         \caption{Input test image}
         \label{subfig:input_quartets}
     \end{subfigure}
     \begin{subfigure}[b]{0.5\textwidth}
        \includegraphics[width=\textwidth]{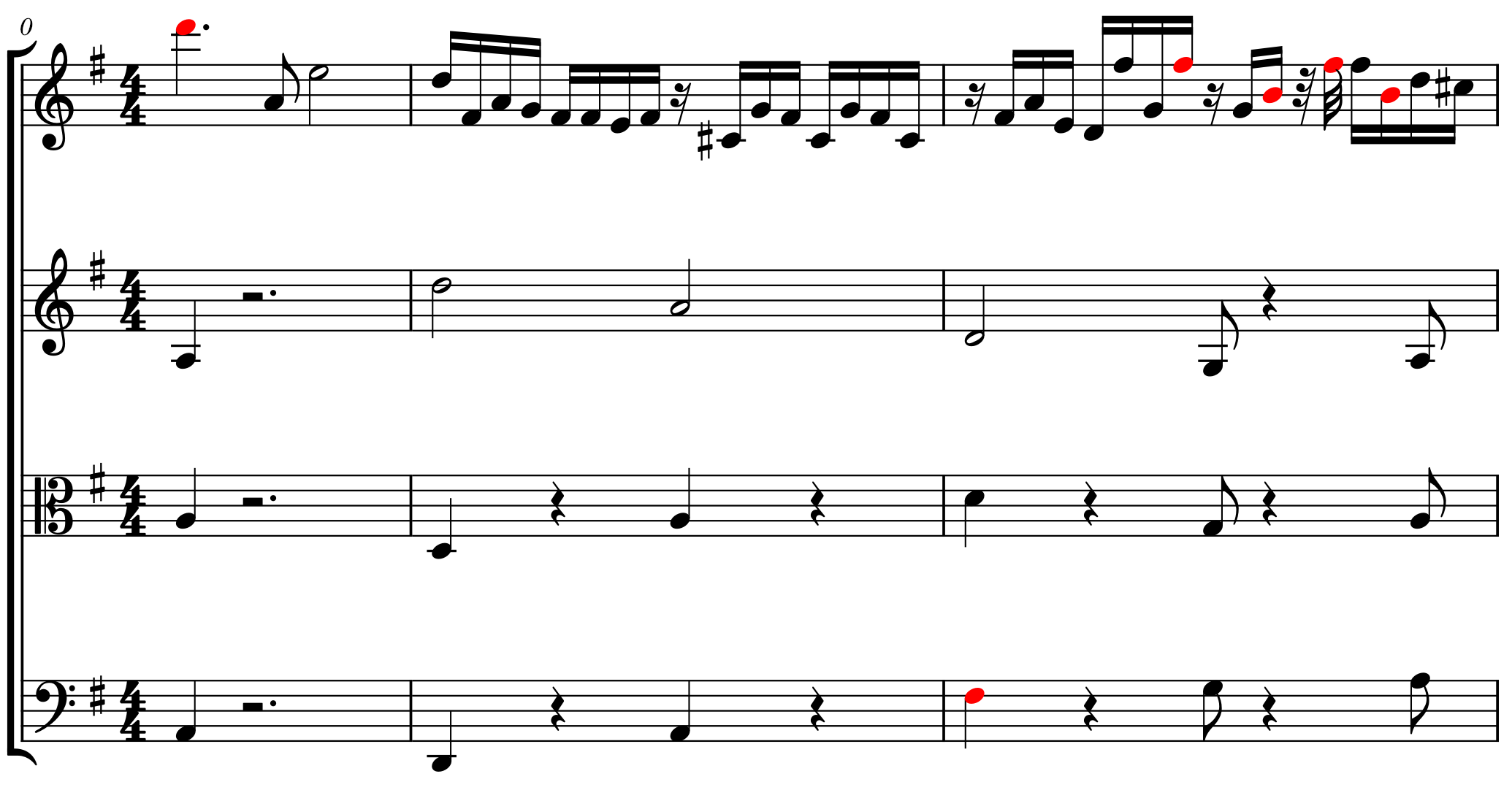}
         \caption{Prediction from the \SMTNXT{}.}
         \label{subfig:output_quartets}
     \end{subfigure}
        \caption{Test example from the Quartets dataset with the errors highlighted. This specific sample attained a \ac{CER} of 1.0\%, a \ac{SER} of 1.8\% and a \ac{LER} of 3.2\%.}
        \label{fig:quartets_test_ex}
\end{figure}

Two visualization errors from both the Camera \GrandStaff{} and Quartets dataset are shown in Figures \ref{fig:grandstaff_test_ex} and \ref{fig:quartets_test_ex}, respectively. The highlighted errors were computed using the musicdiff tool~\cite{Foscarin2019} and visualized with the MuseScore editor\footnote{https://musescore.org/es}.

\begin{figure}[H]
    \centering
    \includegraphics[width=\textwidth]{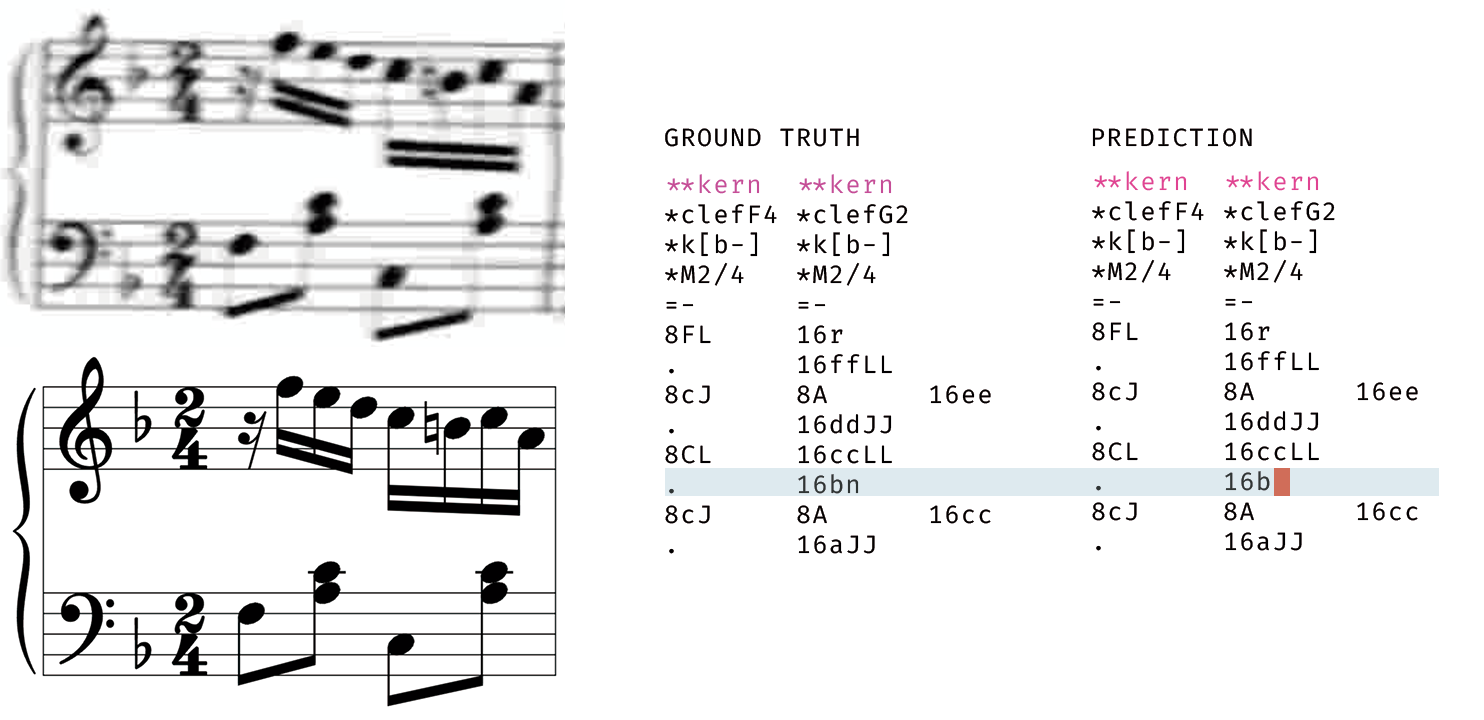}
    \caption{Example of a redundant error in the Camera \GrandStaff{} dataset, in which the ground truth contains redundant annotations that the language model avoids.}
    \label{fig:redundant_case}
\end{figure}

Both examples show that, visually, the model is capable of generating a correct music sequence in terms of syntax, as no bar-completion or time-related errors are found. Indeed, most of the errors that are highlighted in these examples show that most of them are pitch misplacements. This means that, although the model is able to recognize the shape of the notes accurately, it generally fails to predict their position in their staff, or to place accidentals.
Another interesting case is that shown in Figure \ref{fig:redundant_case}. 

In this figure, note that the \ac{SMT} predicted a visually identical excerpt of music. However, when reading the transcription produced in comparison to the ground truth, there is an insertion error---in line 11, which is highlighted in blue and red. Specifically, the model should have predicted the note ‘B’ with a natural accidental (\musNatural{}). This is a specific difficulty that the Humdrum **kern annotation rules has with music renderers. The Humdrum **kern syntax specifies that if a note is affected by an accidental, the annotator must always indicate it, independently of how it is graphically seen. However, music renderers perform the simplifications of Common Western Notation rules when visualizing the music score. This results in cases in which some accidentals are rendered or the music rules are not provided, but these can be found in the ground-truth transcription. 

This is a specific case in music notation in which text-based metrics penalize unfairly, as the piece produced is musically correct and equivalent to the ground truth. It is, therefore, possible that the metrics presented in Table \ref{tab:results}, although necessary to assess document-quality information, are pessimistic figures of merit for the methodological evaluation. Even though no consensus has been reached as to the best way in which to evaluate \ac{OMR}, and that this is still, therefore, an open research question, we corroborate that the results of this paper are robust, usable and musically-accurate, and this is also supported by the information provided in Table \ref{tab:quality_analysis}.

\section{Conclusion}
\label{sec:conclusions}
In this paper, we tackle the challenge of advancing Optical Music Recognition beyond monophonic transcription, which has traditionally involved simplifications or \textit{ad-hoc} adaptations of existing methods. We introduce the \acf{SMT} model, an autoregressive Transformer-based model that utilizes attention computation and language modeling to transcribe input scores into digital music encodings. We evaluate this approach in two polyphonic music scenarios: pianoform and string quartet scores. Our results demonstrate that the SMT model not only effectively transcribes complex musical layouts, but also outperforms current state-of-the-art methods, thus implying a significant advance in \ac{OMR}.

This research also opens avenues for further improvements to \ac{OMR}. Firstly, and as discussed in Section \ref{sec:discussion}, standard \ac{OMR} evaluation methods may overlook musicological interpretations, leading to pessimistic results. Addressing this issue could involve exploring graphic-based output music encodings~\cite{Calvo-Zaragoza2018} or employing additional musicological metrics in order to assess transcription accuracy~\cite{McLeod2018}. The development of segmentation-free full-page transcription methods is another promising direction, as it is not limited by image features or layout constraints. Finally, we propose that the Universal OMR transcription challenge, a desired goal in \ac{OMR} research, could be addressed by profiting from the language modeling capabilities of the \ac{SMT}, thus allowing it to be trained to transcribe music engraved in various manners.
\newpage

\section{Acknowledgements}
This paper is part of the project I+D+i PID2020-118447RA-I00 (MultiScore), funded by MCIN/AEI/10.13039/501100011033. The first author is supported by grants ACIF/2021/356 and CIBEFP/2022/19 from the ``Programa I+D+i de la Generalitat Valenciana''.

\bibliographystyle{splncs04}
\bibliography{bibliography}

\end{document}